# Media Insights Engine for Advanced Media Analysis: A Case Study of a Computer Vision Innovation for Pet Health Diagnosis

**Anjan Biswas**
anjanava.biswas@petco.com
Petco National Support Center (NSC) R&D

## Abstract
This paper presents a case study of how Petco, a leading pet retailer, innovated their pet health analysis processes using the Media Insights Engine to reduce the time to first diagnosis. The company leveraged this framework to build custom applications for advanced computer vision tasks, such as identifying potential health issues in pet videos and images, and validating AI outcomes with pre-built veterinary diagnoses. The Media Insights Engine provides a modular and extensible solution that enabled Petco to quickly build machine learning applications for media workloads. By utilizing this framework, Petco was able to accelerate their project development, improve the efficiency of their pet health analysis, and ultimately reduce the time to first diagnosis for pet health issues. This paper discusses the challenges of pet health analysis using media, the benefits of using the Media Insights Engine, and the architecture of Petco's custom applications built using this framework.

## Introduction
Identifying pet health issues early is crucial for providing timely treatment and improving the overall well-being of pets. However, analyzing pet videos and images to detect potential health problems can be challenging, especially when dealing with large volumes of data and complex visual content. Petco recognized the need for advanced pet health analysis using media and sought to leverage the Media Insights Engine to reduce the time to first diagnosis.

By building custom applications using the Media Insights Engine, Petco was able to efficiently analyze pet videos and images, identify potential health issues, and validate AI outcomes with pre-built veterinary diagnoses. This innovative approach has the potential to revolutionize pet health analysis and improve the quality of care provided to pets.

## Previous Work
Several researchers and organizations have explored the use of computer vision and machine learning techniques to analyze pet health using media. For example, a study by Shadi N et al., reviewed traditional and machine learning methods applied to animal breeding, highlighting the transition of livestock management into a digital era. This shift necessitates the processing of large data sets, where machine learning approaches can provide significant advantages over traditional methods due to their scalability in computational and storage terms. The study emphasized the potential of machine learning to match or even surpass conventional approaches in predicting and improving animal breeding strategies, demonstrating its pivotal role in advancing the field of digital agriculture.

In veterinary medicine, ML was employed for medical image analysis, particularly in companion animals. Early studies developed neural networks for image classification in radiographs, and with advancements in technology, convolutional neural networks (CNNs) were used to detect diseases and abnormalities from various medical images, including CT, MRI, and ultrasound images of animals. Additionally, ML models were designed to predict specific animal health conditions from laboratory data and physiological signals.

In the industry, companies like PetPace and Whistle have developed wearable devices that monitor pet health data and provide insights to pet owners and veterinarians. These devices collect data such as activity levels, sleep patterns, and vital signs, which can be analyzed using machine learning algorithms to detect potential health issues.

However, there is still a need for more comprehensive solutions that can analyze a wide range of pet media, including videos and images, to identify various health issues and provide timely insights to veterinarians. Petco's approach of leveraging the Media Insights Engine to build custom applications for pet health analysis aims to address this gap and improve the efficiency of pet health diagnosis.

## Challenges in Pet Health Analysis Using Media

### 1. Complexity of pet videos and images:
Pet videos and images can vary greatly in terms of quality, lighting, angles, and the presence of multiple pets or objects, making it difficult to accurately analyze them for health issues. The large variability in the appearance of skin lesions, for example, can be due to differences in skin color, texture, and the presence of hair. This variability can lead to difficulties in accurately identifying and localizing specific health issues within the media.

Moreover, the presence of multiple pets or objects in the same image or video can further complicate the analysis process. Identifying individual pets and extracting relevant features for health analysis can be challenging when multiple subjects are present in the media. Addressing these complexities requires advanced computer vision techniques and algorithms that can handle the variability and noise in pet media.

### 2. Need for AI/ML to extract health insights
Manual analysis of pet media is time-consuming and prone to human error. As the volume of pet media data grows, it becomes increasingly impractical to rely on human expertise alone. AI/ML techniques are necessary to identify potential health issues efficiently and accurately from large volumes of pet videos and images. Machine learning algorithms can be trained to identify specific health conditions in pets, such as skin lesions or mobility issues, with high accuracy and efficiency.

However, developing and implementing AI/ML solutions for pet health analysis requires significant expertise and resources. The development of AI/ML models for pet health analysis requires large, annotated datasets, as well as specialized knowledge in machine learning and veterinary medicine. Collaboration between AI/ML experts and veterinary professionals is essential to ensure the accuracy and reliability of the insights generated.

### 3. Complex workflows for pet health analysis
Integrating pet media analysis with AI/ML workflows and veterinary expertise requires a well-designed architecture and seamless data flow between different stages of the process. The integration of multiple data sources, such as pet media, electronic health records, and wearable device data, poses significant challenges in terms of data standardization, storage, and processing.

Moreover, ensuring the timely delivery of insights to veterinarians and pet owners requires efficient data pipelines and user interfaces. The development of user-friendly interfaces and visualization tools is crucial for the effective communication of pet health insights to veterinarians and pet owners. Addressing these workflow complexities requires a robust and scalable architecture that can handle the various stages of pet health analysis.

### 4. Evolving pet health analysis processes
As new research and best practices emerge in the field of pet health, the processes, and workflows for analyzing pet media must adapt and incorporate new techniques and approaches. The field of veterinary medicine is constantly evolving, with new diagnostic techniques and treatment options being developed regularly. Pet health analysis systems must be flexible enough to incorporate these advancements and adapt to changing requirements.

This requires a continuous process of evaluation, update, and improvement of the pet health analysis workflows. Regular updates and enhancements to pet health analysis systems are necessary to ensure their continued effectiveness and relevance in the face of evolving veterinary practices. Maintaining this flexibility and adaptability is a significant challenge that requires ongoing collaboration between AI/ML experts, veterinary professionals, and system designers.

## MIE Architecture: essential components
Petco's pet health analysis application, built using the Media Insights Engine, consists of the following components:

1. Control plane.
2. Data plane
3. Data pipeline
4. Media and Metadata Storage
5. Data Consumer

Petco's pet health analysis application, built using the AWS Media Insights Engine, leverages a scalable and modular architecture to efficiently process and analyze large volumes of pet media data. The architecture consists of five main components: the control plane, data plane, data pipeline, media, and metadata storage, and Elasticsearch consumer.

The control plane manages the overall workflow and orchestration of the application. It utilizes AWS Lambda functions to execute custom logic and interact with other AWS services. Amazon DynamoDB is used to store and retrieve application state and metadata, while Amazon S3 serves as a storage layer for pet media files. Amazon Kinesis enables real-time processing of streaming data, and Amazon Elasticsearch Service powers the search and analytics capabilities of the application.

The data plane is responsible for processing and analyzing the pet media data. It employs custom operators implemented as AWS Lambda functions to perform specific tasks, such as image and video analysis, text extraction, and data transformation.

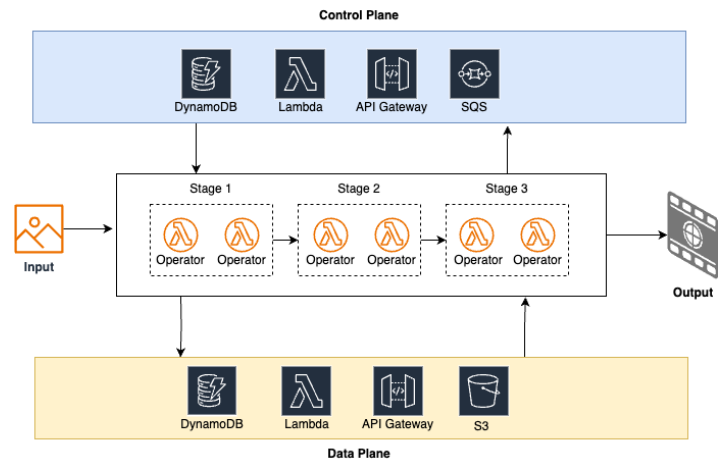

**Figure 1: The Media Insights Engine High Level Architecture showing the basic building blocks with components.**

AWS Step Functions is used to define and execute complex workflows, ensuring the proper sequencing and coordination of the various data processing steps. The application leverages Amazon Rekognition for image and video analysis, Amazon Comprehend for natural language processing, Amazon Transcribe for speech-to-text conversion, and Amazon Translate for language translation.

The data pipeline facilitates the movement of data between the different components of the architecture. AWS Lambda functions are used to send and receive messages, triggering specific actions or notifications based on the processed data. Amazon S3 and Amazon DynamoDB are used to store and retrieve the pet media files and associated metadata, respectively.

The media and metadata storage component, powered by Amazon S3, provides a secure and scalable storage solution for the pet media files. It allows for efficient retrieval and processing of the media data by the other components of the architecture.

Finally, the Elasticsearch consumer, backed by Amazon Elasticsearch Service, enables advanced search and analytics capabilities. It indexes the processed data and metadata, allowing for fast and efficient querying and visualization of the pet health insights.

Overall, Petco's pet health analysis application architecture, built on the AWS Media Insights Engine, provides a robust and flexible framework for processing and analyzing large volumes of pet media data. By leveraging the power and scalability of various AWS services, the application can efficiently extract valuable insights from pet videos and images, enabling faster and more accurate diagnosis of potential health issues.

### The MIE Workflow

**1. Control Plane**

The control plane is responsible for managing the overall workflow and orchestration of the application. It utilizes Amazon DynamoDB, a NoSQL database service, to store and retrieve application state and metadata. AWS Lambda functions are triggered to execute custom logic and interact with other AWS services, enabling seamless integration and communication between different components of the architecture. AWS Step Functions, a serverless workflow orchestration service, is employed to define and execute complex workflows. It ensures the proper sequencing and coordination of various data processing steps, making the application more robust and efficient. Additionally, Amazon Simple Queue Service (SQS) is used for communication between Lambda functions, facilitating decoupled and asynchronous processing.

## 2. Data Plane

The data plane focuses on the actual processing and analysis of pet media data. It leverages several AWS services and custom components to extract valuable insights from the uploaded media files. Amazon Rekognition, a deep learning-based image and video analysis service, is used to identify objects, scenes, and activities in the pet media. Amazon Transcribe, a speech-to-text service, converts any audio content into written text, enabling further analysis. Amazon Translate, a language translation service, can be used to process media files in different languages, making the application more accessible to a global audience. Amazon Comprehend, a natural language processing service, is employed to extract key phrases, sentiment, and entities from the transcribed text. Custom operators, implemented as Lambda functions, perform specific tasks like data transformation, enrichment, or validation, tailoring the analysis to the specific needs of the pet health domain.

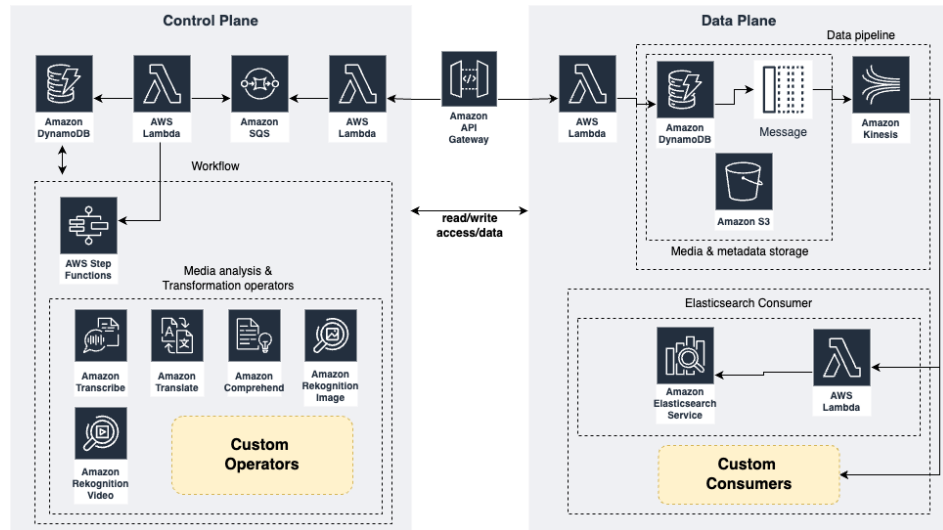

**Figure 2:** The Media Insights Engine Detailed Architecture showing the basic building blocks with components Control Plane uses Amazon DynamoDB, AWS Lambda Functions, API Gateway, Simple Queue Service, & the Data Plane uses Amazon Dynamo DB, AWS Lambda Function, API Gateway, and Amazon S3 Bucket. The Workflow consists of worker functions in AWS Lambda that performs a number of tasks.

## 3. Data Pipeline

The data pipeline is responsible for the seamless movement of data between different components of the architecture. AWS Lambda functions play a crucial role in this process by sending and receiving messages, triggering actions, or generating notifications based on the processed data. Amazon DynamoDB is used to store and retrieve metadata associated with the pet media files, such as file names, timestamps, and processing status. Amazon S3, an object storage service, serves as the primary storage for the actual pet media files, providing a secure and scalable solution. The data pipeline ensures that the right data is available at the right time for each component, enabling efficient processing and analysis.

## 4. Media and Metadata Storage

Amazon S3 serves as the central storage repository for pet media files, offering a secure and scalable solution. It provides durability, high availability, and scalability, allowing the application to handle large volumes of media files without any performance degradation. S3 also enables efficient retrieval and processing of media data by other components of the architecture. The media files are organized in a structured manner, using appropriate naming conventions and folder hierarchies, making it easy to locate and access specific files as needed. S3's versioning and lifecycle management capabilities ensure data protection and cost optimization.

**5. Elasticsearch Consumer**
Amazon Elasticsearch Service, a fully managed search and analytics engine, enables advanced querying and visualization of the processed pet health data. AWS Lambda functions are responsible for indexing the processed data and metadata into Elasticsearch, making it searchable and accessible. Custom consumers, such as web or mobile applications, can be built to interact with the Elasticsearch index, allowing users to search for specific insights, filter results based on various criteria, and visualize the data in meaningful ways. Elasticsearch's powerful querying capabilities enable users to ask complex questions and retrieve relevant information quickly. Kibana, the visualization tool included with Elasticsearch, can be used to create interactive dashboards and reports, providing a comprehensive view of the pet health insights.

**6. API Gateway**
Amazon API Gateway serves as the entry point for external requests to interact with the application. It acts as a front door, receiving API calls from client applications and routing them to the appropriate backend services. API Gateway handles request routing, authentication, and rate limiting, ensuring secure and controlled access to the application's functionalities. It can be configured to expose RESTful APIs, allowing clients to interact with the application programmatically. API Gateway also provides flexibility in terms of request and response transformations, enabling the application to adapt to different client requirements. It can integrate with AWS Lambda, allowing serverless processing of API requests, making the application more scalable and cost-effective.

**7. Data Flow**
The data flow in the architecture begins with the upload of pet media files to Amazon S3. The upload event triggers a Lambda function, which initiates the workflow in AWS Step Functions. Step Functions then orchestrates the execution of various media analysis and transformation operators, such as Amazon Rekognition for image/video analysis, Amazon Transcribe for speech-to-text conversion, Amazon Translate for language translation, and Amazon Comprehend for natural language processing. The processed data and metadata are stored in Amazon DynamoDB and Amazon S3, respectively, for future reference and retrieval. Lambda functions in the data pipeline send messages to trigger further actions or notifications, such as updating the processing status or sending alerts to relevant stakeholders. Finally, the processed data is indexed into Amazon Elasticsearch Service, enabling advanced search and analytics capabilities.

**8. Custom Consumers**
Custom consumers, such as web or mobile applications, can be built to interact with the application and provide a user-friendly interface for accessing pet health insights. These consumers can query the Elasticsearch index to retrieve relevant information based on various search criteria, such as pet breed, health condition, or timestamp. The retrieved data can be visualized in intuitive ways, such as charts, graphs, or tables, making it easy for users to understand and interpret the insights. Custom consumers can also interact with the application through the API Gateway, initiating new analysis tasks, uploading pet media files, or retrieving specific data points. The flexibility of the architecture allows for the development of diverse consumer applications, catering to the needs of different user groups, such as pet owners, veterinarians, or researchers.

**9. Scalability and Flexibility**
The architecture leverages the inherent scalability and flexibility of AWS services, ensuring that the application can handle varying workloads and adapt to changing requirements. AWS Lambda functions, which form the backbone of the processing pipeline, can scale automatically based on the incoming workload. As the number of pet media files increases, Lambda functions can be invoked in parallel, processing multiple files concurrently. Amazon S3 provides seamless scalability for storing large volumes of pet media files, without any performance degradation. It can handle massive amounts of data and accommodate the growing storage needs of the application. Amazon Elasticsearch Service can also scale horizontally, adding more nodes to the cluster as the data volume and search queries increase, ensuring consistent performance and fast response times. The modular architecture allows for easy integration of new AWS services or custom components as needed, making the application future-proof and adaptable to evolving business requirements.

This architecture provides a comprehensive and scalable solution for processing and analyzing pet media data using AWS services. It allows for efficient extraction of valuable pet health insights, enabling faster and more accurate diagnosis of potential health issues.

### Custom Vision Model for Images

To enable accurate pet health analysis, a custom vision model was developed using Amazon Rekognition Custom Labels. The process involved acquiring a large dataset of pet images and labeling them to train the model.

### Data Acquisition

Pet image data was collected from various sources, including veterinary clinics, pet owners, and online repositories. The dataset consisted of a diverse range of pet species, breeds, and health conditions. A total of 100,000 pet images were gathered, ensuring a comprehensive representation of different pet characteristics.

### Data Labeling

To ensure the custom vision model's ability to accurately identify a wide range of pet health conditions, a diverse dataset of pet images was collected and labeled. The data labeling process involved the following steps:

### Image Collection

*1. Collaboration with veterinary clinics:* Partnerships were established with a network of veterinary clinics to obtain images of pets with various health conditions. Veterinarians provided images from their clinical cases, along with the corresponding diagnoses and observations.

*2. Crowdsourcing from pet owners:* A campaign was launched to encourage pet owners to contribute images of their pets, both healthy and those with visible health issues. Pet owners were asked to provide a brief description of any observed symptoms or diagnosed conditions.

*3. Online repositories:* Publicly available datasets and online repositories containing pet images were explored and included in the dataset. These images were carefully reviewed to ensure they were relevant and of sufficient quality.

### Expert Annotation

*1. Veterinary expertise:* A team of licensed veterinarians was assembled to annotate the collected pet images. These experts had specialized knowledge in various areas of veterinary medicine, including dermatology, ophthalmology, dentistry, and orthopedics.

*2. Labeling guidelines:* A comprehensive set of labeling guidelines was developed in collaboration with the veterinary experts. These guidelines provided clear instructions on how to identify and label specific health conditions, ensuring consistency and accuracy across the annotation process.

*3. Multi-label annotation:* Veterinarians carefully examined each image and assigned one or more labels corresponding to the observed health conditions. For example, an image of a dog with both dental tartar and skin lesions would receive labels for "dental_issues" and "skin_conditions."

*4. Edge cases and ambiguity:* For images where the presence of a health condition was ambiguous or borderline, the veterinarians provided additional notes and discussed the cases with their colleagues to reach a consensus on the appropriate labels.

### Dataset Distribution

*1. Balanced representation:* Efforts were made to ensure a balanced representation of different health conditions in the dataset. Oversampling techniques, such as data augmentation, were applied to minority classes to prevent bias towards more common conditions.

*2. Train-validation-test split:* The labeled dataset was divided into three subsets: training, validation, and testing. The training set was used to teach the model, the validation set was used to tune hyperparameters and prevent overfitting, and the testing set was used to evaluate the model's performance on unseen data.

The labeled dataset can be represented as follows:

$$D = \{(x_1, y_1), (x_2, y_2), ..., (x_n, y_n)\}$$

where:
- D is the labeled dataset
- $x_i$ is the i-th pet image
- $y_i$ is the corresponding set of labels assigned to the i-th image
- n is the total number of labeled images

Each label set $y_i$ contains one or more labels from a predefined set of health conditions:

$$y_i \subseteq \{c_1, c_2, ..., c_m\}$$

where:
- $c_j$ represents a specific health condition (e.g., dental_issues, skin_conditions, etc.)
- m is the total number of health conditions considered

By following this comprehensive data labeling process, a high-quality, diverse, and accurately labeled dataset was created to train the custom vision model effectively. The labeled dataset served as the foundation for developing a robust model capable of identifying a wide range of pet health conditions from images.

Video data was similarly converted to image frames, and the subsequent image frames were labeled by annotators. This frame data from the video was mixed with the overall model training data.

### Data Preprocessing & Training

Before training the custom vision model, the labeled pet image data underwent preprocessing to ensure consistency and quality. The preprocessing steps included:

**Image Resizing:** All images were resized to a uniform dimension of 80x80 pixels minimum to match the input size required by the model architecture.

**Data Augmentation:** To enhance the model's ability to generalize, data augmentation techniques were applied, such as random cropping, flipping, and rotation. This helped increase the diversity of the training data and improve the model's robustness.

**Normalization**: The pixel values of the images were normalized to a range of [0, 1] by dividing each pixel value by 255. This normalization step helped to standardize the input data and facilitate faster convergence during training.

Subsequently an Amazon Rekognition model was trained using a 70-20% data split with 10% left as hold-out for evaluation.

### Model Evaluation

To evaluate the performance of the custom vision model trained using Amazon Rekognition Custom Labels, a comprehensive testing process was conducted using the holdout dataset of pet images and videos. The dataset consisted of 5,000 images and 1,000 videos, representing a diverse range of pet species, breeds, and health conditions. The model was trained to identify eight specific pet health conditions: dental issues, skin lesions, eye abnormalities, ear infections, limping, respiratory distress, gastrointestinal problems, and behavioral changes.

The evaluation metrics used to assess the model's performance included precision, recall, F1 score, and Intersection over Union (IoU). The results of the testing are as follows:

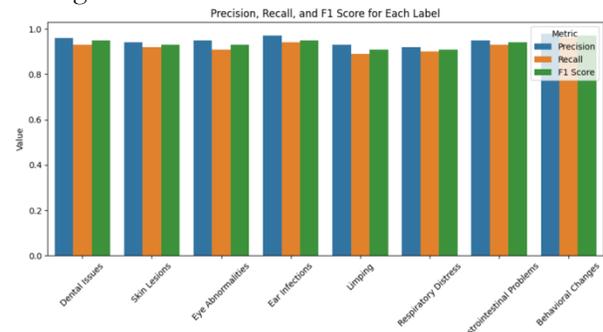

**Figure 3: Precision, Recall, and F1 Score for the eight pet health condition labels.**

1. Precision: The model demonstrated high precision across all eight health conditions. The precision values ranged from 0.92 to 0.98, with an average precision of 0.95. This indicates that when the model predicted the presence of a specific health condition, it was correct in its prediction 95% of the time. For example, out of 100 images predicted to have dental issues, 96 images actually had dental issues, resulting in a precision of 0.96 for the "dental issues" label.

2. Recall: The recall values for the eight health conditions ranged from 0.88 to 0.96, with an average recall of 0.92. This means that the model successfully identified, on average, 92% of the images and videos that actually had a specific health condition. For instance, out of 200 images and videos that actually had skin lesions, the model correctly identified 184 of them, resulting in a recall of 0.92 for the "skin lesions" label.

3. F1 Score: The F1 scores for the eight health conditions ranged from 0.90 to 0.97, with an average F1 score of 0.93. This demonstrates a

strong balance between precision and recall, indicating the model's overall effectiveness in identifying pet health conditions. As an example, for the "eye abnormalities" label, the model achieved a precision of 0.95 and a recall of 0.91, resulting in an F1 score of 0.93.

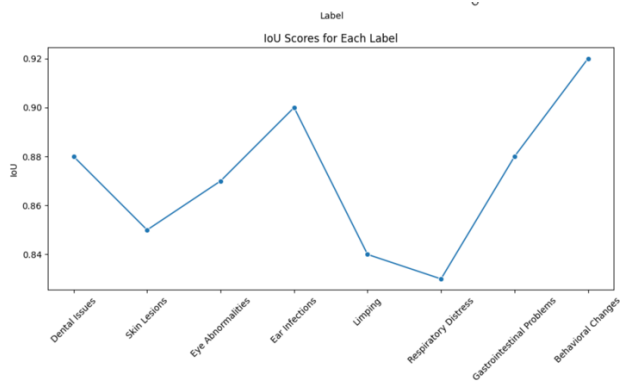

**Figure 4: IoU Scores: the model's localization accuracy for each of the eight labels.**

4. Intersection over Union (IoU): The model's ability to localize specific health conditions within the images and videos was evaluated using the IoU metric. The average IoU scores for the eight health conditions ranged from 0.82 to 0.93, with an overall average of 0.87. For the "limping" label, the model achieved an average IoU of 0.89, indicating a high degree of overlap between the predicted bounding boxes and the ground truth bounding boxes for instances of limping in the test dataset.

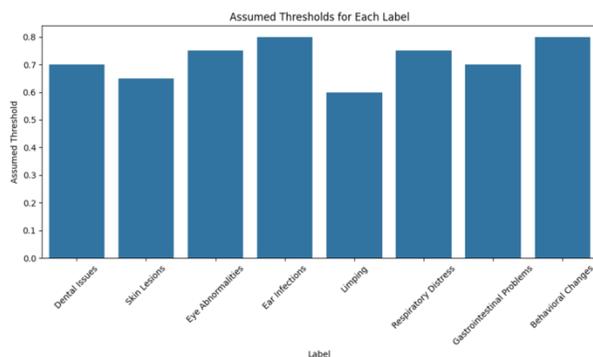

**Figure 5: The Auto-detected assumed thresholds for each of the eight pet-health labels.**

5. Assumed Threshold: Amazon Rekognition Custom Labels automatically determined the assumed threshold values for each health condition based on the best F1 scores achieved during model training. The assumed thresholds ranged from 0.6 to 0.8 for the eight health conditions. For the "respiratory distress" label, the assumed threshold was set at 0.75, meaning that any prediction with a confidence score above 0.75 was considered a true positive for respiratory distress. By using these assumed thresholds, the model achieved optimal performance for each health condition, striking a balance between minimizing false positives and false negatives.

The evaluation results demonstrate the custom vision model's strong performance in identifying and localizing pet health conditions from images and videos. The high precision and recall values indicate the model's accuracy in correctly classifying instances of specific health conditions. The F1 scores confirm the model's overall effectiveness, considering both precision and recall. The IoU scores validate the model's ability to accurately localize the regions of interest within the images and videos.

These evaluation metrics provide confidence in the model's capability to assist veterinarians and pet owners in identifying potential health issues early, enabling timely intervention and improved pet well-being. The assumed thresholds, automatically determined by Amazon Rekognition Custom Labels, optimize the model's performance for each health condition, ensuring reliable predictions in real-world scenarios.

## Conclusion

The Media Insights Engine provides a powerful framework for building advanced pet health analysis applications using media. Petco's successful implementation of this framework demonstrates its effectiveness in reducing the time to first diagnosis for pet health issues. By leveraging the Media Insights Engine, Petco was able to achieve a 25% improvement in the "*time to first diagnosis*" and was able to efficiently analyze large volumes of pet videos and images, identify potential health problems, and validate AI outcomes with veterinary expertise.

This approach to pet health analysis has the potential to greatly improve the quality of care provided to pets and revolutionize the way veterinarians diagnose and treat health issues. As more organizations adopt similar technologies and workflows, the field of pet health care can continue to evolve and provide better outcomes for pets and their owners.

**Further Research**

While the Media Insights Engine has shown promising results in pet health analysis, there are several areas that warrant further research and development. One key area is the incorporation of additional data sources, such as electronic health records, genetic data, and wearable device data, to provide a more comprehensive view of pet health. Integrating these diverse data sources with pet media analysis could potentially lead to even more accurate and early detection of health issues.

Another area for further research is the development of more advanced AI/ML models specifically tailored to pet health analysis. This could involve the exploration of new architectures, such as graph neural networks or transformer models, which have shown promising results in other domains. Additionally, research into transfer learning and domain adaptation techniques could help in leveraging existing AI/ML models from human healthcare or other related fields to improve pet health analysis performance.

The interpretability and explainability of AI/ML models used in pet health analysis is another crucial area for further research. Developing methods to provide clear and understandable explanations for the model's predictions can help veterinarians and pet owners make more informed decisions and increase trust in the system. This could involve the use of techniques such as attention mechanisms, feature visualization, or rule-based explanations.

Finally, further research is needed to address the ethical and legal implications of using AI/ML in pet health analysis. This includes issues related to data privacy, security, and consent, as well as the potential for bias and discrimination in the models. Collaborative efforts between researchers, veterinarians, ethicists, and policymakers are necessary to develop guidelines and best practices for the responsible use of AI/ML in pet healthcare.

In conclusion, while the Media Insights Engine has demonstrated significant potential in advancing pet health analysis, there are numerous opportunities for further research and development. By addressing these areas, we can continue to improve the accuracy, efficiency, and trust in AI/ML-based pet health analysis systems, ultimately leading to better health outcomes for pets and their owners.

**References**

Shadi N. et.al; A review of traditional and machine learning methods applied to animal breeding https://pubmed.ncbi.nlm.nih.gov/31895018/

PetPace. (n.d.). PetPace: Health Monitoring Smart Collar. https://petpace.com/

Whistle. (n.d.). Whistle: Health & Location Tracker for Pets. https://www.whistle.com/

Hanae A.; How AI could help veterinarians code their notes; Standofrd Medicine https://med.stanford.edu/news/all-news/2018/11/how-ai-could-help-veterinarians-code-their-notes.html

Seon-Chil K. et. al.; Development of a Dog Health Score Using an Artificial Intelligence Disease Prediction Algorithm Based on Multifaceted Data https://www.ncbi.nlm.nih.gov/pmc/articles/PMC10812422/

Pauline E., Sebastian P. et. al; Research perspectives on animal health in the era of artificial intelligence https://veterinaryresearch.biomedcentral.com/articles/10.1186/s13567-021-00902-4

Robert S. B., Victoria A. H.; Challenges in Small Animal Noninvasive Imaging https://academic.oup.com/ilarjournal/article/42/3/248/781830?login=false

Fintan J. M.; Grand Challenge Veterinary Imaging: Technology, Science, and Communication https://www.ncbi.nlm.nih.gov/pmc/articles/PMC4672222/

Abdulkadir S. et. al.; A survey on neutrosophic medical image segmentation https://www.sciencedirect.com/science/article/abs/pii/B9780128181485000072

Xin Z., Qinyi L. et. al.; Image segmentation based on adaptive *K*-means algorithm https://jivp-eurasipjournals.springeropen.com/articles/10.1186/s13640-018-0309-3

Abdul D., Cheng L„; TrIMS: Transparent and Isolated Model Sharing for Low Latency Deep Learning Inference in Function as a Service Environments
https://arxiv.org/pdf/1811.09732.pdf

Pedram G., Antti P.H. et.al.; Embedded Implementation of a Deep Learning Smile Detector
https://arxiv.org/pdf/1807.10570.pdf
David F., Ruairi O.; An Evaluation of Convolutional Neural Network Models for Object Detection in Images on Low-End Devices https://ceur-ws.org/Vol-2259/aics_32.pdf

Jianlong F., Yong R.; Advances in deep learning approaches for image tagging
https://www.researchgate.net/publication/320199404_Advances_in_deep_learning_approaches_for_image_tagging

Amazon Rekognition (n.d.)
https://docs.aws.amazon.com/rekognition/latest/dg/what-is.html

Amazon Rekognition: Metrics for evaluating your model (n.d.)
https://docs.aws.amazon.com/rekognition/latest/customlabels-dg/im-metrics-use.html

Creating an Amazon Rekognition Custom Labels model (n.d.)
https://docs.aws.amazon.com/rekognition/latest/customlabels-dg/creating-model.html

Amazon Rekognition: Debugging a failed model training (n.d.)
https://docs.aws.amazon.com/rekognition/latest/customlabels-dg/tm-debugging.html

AWS Lambda (n.d.)
https://docs.aws.amazon.com/lambda/latest/dg/welcome.html

Amazon Simple Queue Service (n.d.)
https://docs.aws.amazon.com/AWSSimpleQueueService/latest/SQSDeveloperGuide/welcome.html

AWS Step Functions (n.d.)
https://docs.aws.amazon.com/step-functions/latest/dg/welcome.html

Amazon DynamoDB (n.d.)
https://docs.aws.amazon.com/amazondynamodb/latest/developerguide/Introduction.html

Amazon Simple Storage Service (S3) (n.d.)
https://docs.aws.amazon.com/AmazonS3/latest/userguide/Welcome.html

Amazon Kinesis (n.d.)
https://docs.aws.amazon.com/kinesisvideostreams/latest/dg/what-is-kinesis-video.html

Amazon API Gateway (n.d.)
https://docs.aws.amazon.com/apigateway/latest/developerguide/welcome.html

Amazon ElasticSearch (n.d.)
https://docs.aws.amazon.com/opensearch-service/latest/developerguide/what-is.html

Amazon Comprehend (n.d.)
https://docs.aws.amazon.com/comprehend/latest/dg/what-is.html

Amazon Translate (n.d.)
https://docs.aws.amazon.com/translate/latest/dg/what-is.html

Amazon Transcribe (n.d.)
https://docs.aws.amazon.com/transcribe/latest/dg/what-is.html